\documentclass[conference]{IEEEtran}

\IEEEoverridecommandlockouts                              

\usepackage[utf8]{inputenc} 
\usepackage{amsmath,amsfonts,amssymb,amsthm} 
\usepackage{graphicx} 
\usepackage[dvipsnames]{xcolor} 
\usepackage{flushend} 
\usepackage{setspace} 
\usepackage{caption} 
\usepackage{subcaption} 
\usepackage{comment} 
\usepackage{tikz} 
\usepackage{bm} 
\usepackage{xspace} 
\usepackage{algpseudocode} 
\usepackage{algorithm} 
\usepackage{enumerate} 
\usepackage{listings} 
\usepackage{float} 
\usepackage{csvsimple} 
\usepackage[export]{adjustbox} 
\usepackage{float}
\newfloat{algorithm}{t}{lop}
\usepackage{optidef}

\usepackage[numbers,sort&compress]{natbib} 
\usepackage{amsfonts} 
\usepackage{amsopn} 
\usepackage{breqn} 
\usepackage{hyperref} 
\usepackage{epstopdf} 
\usepackage{lipsum} 
\usepackage{epsfig} 
\usepackage{enumitem}
\setlist{nolistsep}
\usepackage{soul} 
\usepackage{hyperref}
\hypersetup{
    colorlinks=true,
    linkcolor=MidnightBlue,   
    citecolor=MidnightBlue,   
    urlcolor=MidnightBlue     
}

\DeclareGraphicsExtensions{.pdf,.eps,.png,.jpg,.mps} 

\begin{document}

\title{Neural Process-Based Reactive Controller for Autonomous Racing\\
}

\author{\IEEEauthorblockN{Devin Hunter}
\IEEEauthorblockA{Department of Electrical Engineering \\
\textit{University of Central Florida}\\
Orlando, United States \\
de700090@ucf.edu}
\and
\IEEEauthorblockN{Chinwendu Enyioha}
\IEEEauthorblockA{Department of Electrical Engineering \\
\textit{University of Central Florida}\\
Orlando, United States \\
cenyioha@ucf.edu}
}

\maketitle

\begin{abstract}
Attention-based neural architectures have become central to state-of-the-art methods in real-time nonlinear control. As these data-driven models continue to be integrated into increasingly safety-critical domains, ensuring statistically grounded and provably safe decision-making becomes essential. This paper introduces a novel reactive control framework for gap-based navigation using the Attentive Neural Process (AttNP) and a physics-informed extension, the PI-AttNP. Both models are evaluated in a simulated F1TENTH-style Ackermann steering racecar environment, chosen as a fast-paced proxy for safety-critical autonomous driving scenarios. The PI-AttNP augments the AttNP architecture with approximate model-based priors to inject physical inductive bias, enabling faster convergence and improved prediction accuracy suited for real-time control. To further ensure safety, we derive and implement a control barrier function (CBF)-based filtering mechanism that analytically enforces collision avoidance constraints. This CBF formulation is fully compatible with the learned AttNP controller and generalizes across a wide range of racing scenarios, providing a lightweight and certifiable safety layer. Our results demonstrate competitive closed-loop performance while ensuring real-time constraint satisfaction.
\end{abstract}


\section{Introduction \& Background}

Reactive navigation remains a cornerstone of autonomous mobile robotics, particularly in constrained and dynamic environments where fast decision-making is critical. Classical methods such as Follow-The-Gap (FTG) controllers \cite{sezer2012novel}, Vector Field Histograms \cite{borenstein1991vector}, and artificial potential field approaches \cite{khatib1986real} have long served as effective local planners due to their computational efficiency and deterministic structure. However, these techniques often suffer from significant limitations: they are prone to local minima, sensitive to sensor noise, and frequently generate suboptimal or unsafe trajectories when deployed in real-world settings \cite{koren1991potential}.

In parallel, the rise of data-driven control policies—driven by advances in machine learning—has enabled substantial progress in end-to-end autonomous driving. Techniques such as behavioral cloning, deep imitation learning, and reinforcement learning have demonstrated the potential to generalize complex behaviors from raw sensory input \cite{bojarski2016end}. These models offer promising adaptability and scalability but face critical challenges: they require vast quantities of training data, often lack uncertainty estimates that are critical for safety \cite{kendall2017uncertainties}, and tend to overfit to specific environments, resulting in poor generalization performance \cite{kiran2021deep}.

To address these safety and stability shortcomings, recent work has explored augmenting learning-based control architectures with principled control-theoretic frameworks. Methods such as Control Barrier Functions (CBFs) \cite{ames2016control}, Control Lyapunov Functions (CLFs), and contraction metrics \cite{sun2021learning} provide structured guarantees on safe or stable operation. For instance, CBFs can enforce forward-invariant safe sets to ensure the system remains in collision-free regions of the state space, while CLFs promote convergence to goal states. Contraction-based methods offer stronger guarantees by enforcing exponential convergence of trajectories, yielding robustness to modeling error and disturbances. These approaches have increasingly been integrated into model-based reinforcement learning and imitation learning pipelines to enforce hard safety constraints even in the presence of learned uncertainty \cite{berkenkamp2017safe}.

Neural Processes (NPs) offer a compelling alternative in this space by blending the function-approximation power of deep learning with the principled uncertainty modeling of Gaussian processes. As meta-learned conditional stochastic processes, NPs can rapidly infer a distribution over functions from a few context points and produce coherent predictions with quantifiable uncertainty \cite{kim2019attentive,dubois2020npf}. This ability enables fast adaptation to novel inputs with minimal data, while providing confidence estimates that are useful for safety. When used in reactive control, NPs present the opportunity to construct controllers that not only learn quickly and generalize better but also operate with interpretable probabilistic latent structures.

This work proposes a novel NP-based FTG controller that fuses the inductive bias of FTG’s gap-centric reasoning with the representational power of Neural Processes. By using an Attentive Neural Process (AttNP) architecture \cite{kim2019attentive} and a physics-informed variant (PI-AttNP), the model efficiently captures prior geometric structure in the control space while allowing for rapid convergence and low prediction error. The proposed controller demonstrates strong performance in high-speed autonomous navigation scenarios (such as the F1Tenth racing platform), effectively addressing key shortcomings of both classical heuristics and purely learned policies. 

The remainder of this paper is structured as follows. In Section \ref{sec:problem-statement}, we formally state the learning problem tackled by our NP-based approach in the autonomous racing application. Sections \ref{sec:method} and \ref{sec:results} detail the proposed methodology and experimental results, respectively. We conclude with final remarks and avenues for future work in Section \ref{sec:conclude+future}.

\section{Problem Statement} \label{sec:problem-statement}

In this study, we desire to control an Ackermann-style vehicle that can be represented as having the following nonlinear, control-affine dynamics:
\begin{equation}
    \dot{x} = f(x)+g(x)u,
\end{equation}
where $x\in\mathbb{R}^{n}$ represents the system states, $u\in\mathbb{R}^{m} $ represents the system controls, $f(x):\mathbb{R}^{n}\rightarrow\mathbb{R}^{n}$ represents the drift term, and $g(x)$ is the control-related term that is affine with respect to control $u$. Based on these provided dynamics, the problem of interest for our PI-AttNP and other deep learning-based agents is to learn an \emph{end-to-end} follow-the-gap (FTG)-based navigation policy $\pi_{\text{FTG}}$ via an imitation learning methodology where previously-recorded driving data from several racetracks are utilized to train these models. Explicitly, the FTG policy $\pi_{\text{FTG}}$ and learned policy $\pi_{\Gamma}$ with learned parameters $\Gamma$ are modeled as the following:
\begin{equation}
    \begin{aligned}
    \pi_{\text{FTG}}(\psi_{k}) &\rightarrow\delta_{k+1},v_{k+1} \\
    \pi_{\Gamma}(\psi_{k},v_{k},\omega_{k}) &\rightarrow \delta_{k+1}, v_{k+1}
    \end{aligned}
\end{equation}
where $\psi_{k}\in \mathbb{R}^{o}$ represents the $o-$length collection of LiDAR distance scans collected at time step $k$, $v_{k}\in\mathbb{R}$ represents the current observed linear velocity of the car, and $\omega_{k}\in\mathbb{R}$ represents the current observed angular velocity of the car. Both the FTG and learned policies are expected to output the next commanded steering angle $\delta_{k+1}$ and linear velocity $v_{k+1}$ at each time step. However, to account for the unpredictable nature of deep learning-based controllers \cite{ames2016control,sun2021learning,berkenkamp2017safe}, we augment model predictions with an analytically derived control barrier function (CBF) to capture theoretical guarantees of safety from a specified safety function $h(\cdot)$, which prevents the model from generating unsafe commands. To see more details regarding our CBF design and implementation, refer to subsection \ref{subsec:CBF-derivation} within the Methodology. 

\section{Methodology}\label{sec:method}
The objective is to estimate the commanded steering angle $\delta_{k+1}$ and forward, linear velocity $v_{k+1}$ using LiDAR scan data $\psi_k$ and velocity information $v_{k},\omega_{k}$ at time step $k$. This is traditionally done using the follow-the-gap (FTG)-based navigation policy $\pi_{\text{FTG}}(\psi_{k})\rightarrow\delta_{k+1},v_{k+1}$ which only utilizes LiDAR scans alone in its policy definition. In this study, $\pi_{\text{FTG}}$ is used as an expert policy in an imitation learning paradigm, which is learned explicitly by both the attentive neural process and a novel variation - the physics-informed attentive neural process (PI-AttNP). To simplify the control problem, we note that all model variations of the learned policy $\pi_{\Gamma}(\psi_{k},v_{k},\omega_{k})$ and expert FTG policy $\pi_{\text{FTG}}(\psi_{k})$ directly generate steering angle predictions $\delta_{k+1}$ while linear velocity command $v_{k+1}$ is derived by a hand-crafted heuristic based on $\delta_{k+1}$. This heuristic encourages fast speeds in small steering angles and slower speeds as the steering angle increases. See subsection \ref{subsec:speed-heuristic} to obtain more information regarding the speed heuristic. To understand the structure of the PI-AttNP, observe the computational diagram of the original AttNP with the inclusion of $g(\cdot)$ (i.e. the PI-AttNP) found in Figure \ref{fig:attnp-forward}.

\subsection{Model Forward Pass Discussion}

\begin{figure}
    \centering
    \includegraphics[scale=0.27]{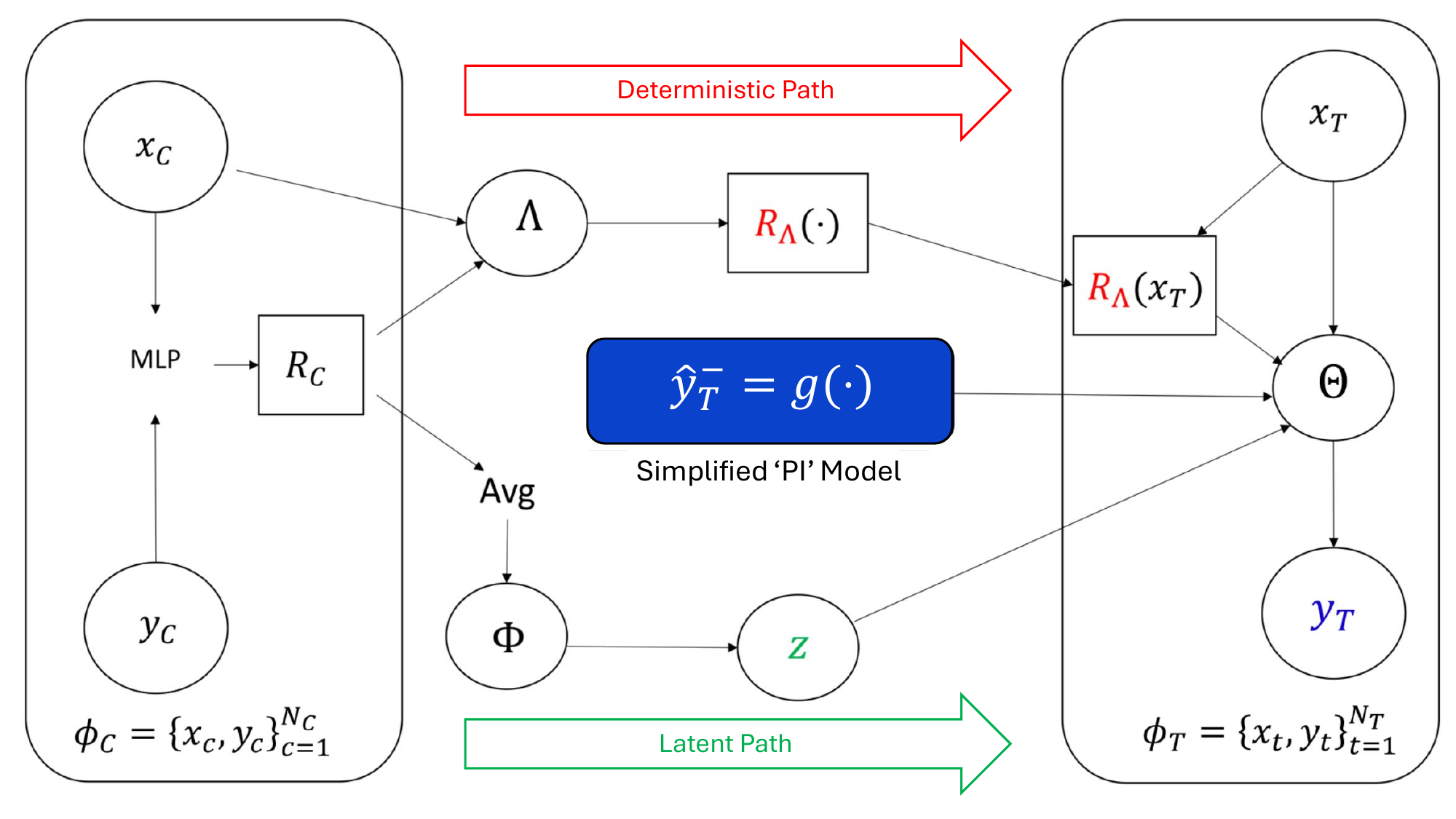}
    \caption{Computational Graph of PI-AttNP's Forward Pass}
    \label{fig:attnp-forward}
\end{figure}

As can be observed in Figure \ref{fig:attnp-forward}, the PI-AttNP maps contextual sets $\phi_{C} = \left\{x_{c}, y_{c}\right\}_{c=1}^{N_{C}}$ to a global representation $R_{C}$ through an initial embedding via a Multi-Layer Perceptron (MLP). For this FTG application, it is noted that $x_{C}$ can represent previous/current dynamic information propagating from the racing vehicle, such as previous LiDAR scans $\psi_{k-1}$ and vehicle dynamics $v_{k-1},\omega_{k-1}$ where $\omega_{k-1}$ is the previous observed yaw angular velocity. From there, the obvious choice for contextual output $y_{C}$ consists of the current commanded steering angle $\delta_k$. The data from the contextual set is used to infer desired next control actions within the target set $\phi_T$.

After dynamic set $\phi_{C}$ is transformed into $R_{C}$, it is sent through split deterministic and latent paths within the forward pass of the network. In the latent path, $R_{C}$ is, first, sent through an averaging operation $\textbf{Avg}$ across individual representations $r_{c}\in R_{C}$ derived from each $\{x_{c},y_{c}\}\in \phi_{C}$. This order-invariant computation is performed to ensure that $\phi_{C}$ maintains the required exchangability condition needed to parameterize the latent variable $z$. After this step, averaged $\phi_{C}$ is used as input to a latent encoder MLP $\Phi(\cdot)$ that is used in the variational approximation of prior distribution $p(z|\phi_{C})$ and posterior distribution $p(z|\phi_{T})$; where $\phi_{T} = \left\{x_{t}, y_{t}\right\}_{t=1}^{N_{T}}$ denotes the target set. Within this FTG application, it is noted that similar to context input $x_C$, target input $x_{T}$ represents the most recent LiDAR scans $\psi_k$ and vehicle dynamics $v_{k},\omega_{k}$; and about target output $y_{T}$, this represents the next commanded steering angle $\delta_{k+1}$.

Through the deterministic path, $R_{C}$ is transformed into a \emph{context-aware} attention matrix $R_{\Lambda}(x_{C},y_{C},x_{T})$ using a deterministic encoder MLP, $\Lambda(\cdot)$, which possesses a cross attention mechanism. Our AttNP approach utilizes multi-headed attention within the cross attention mechanism found in $\Lambda$, which computes parallel scaled dot-products of contextual inputs $x_{C}$ (keys), target inputs $x_{T}$ (queries), and global representation $R_{C}$ (values). Our approach builds on the results in \cite{kim2019attentive}, where the benefits of utilizing eight parallel heads to attain maximum learning efficiency were highlighted. Furthermore, our approach uses self-attention between context pairs $\{x_{C},y_{C}\}$ to better model how aspects of environment scans $\psi_{k-1}$ and velocities $v_{k-1},\omega_{k-1}$ affect the synthesis of previous steering command $\delta_k$ from a transitional dynamics perspective to assist in predicting $\delta_{k+1}$.

Our proposed architecture utilizes a decoder MLP, $\Theta(\cdot)$, to map a set of queried inputs $x_{T}$, attention matrix $R_{\Lambda}$, latent sample $z\sim\mathcal{Z}$, and aprior control estimate $\hat{y}_{T}^{-}\in\mathbb{R}^{m}$ from $g(\cdot)$ to a predictive distribution over queried outputs $p_{\Theta}(y_T | x_{T}, R_{\Lambda},z,\hat{y}_{T}^{-})$. To observe how control prior $g(\cdot)$ is incorporated into the NP's predictive distribution within this FTG application, see subsection \ref{subsec:model-used-for-PINP} in the Results.

It should also be noted that given the roots of PI-AttNP in probabilistic modeling from GPs \cite{dubois2020npf}, we can define context and target set sizes $N_{C},N_{T}$ to be arbitrarily set. Within our two-step control application, we set $N_{C}=N_{T}=1$ to ensure that only sensor measurements within time window $t=k-1,k$ are used to infer the steering angle $\delta_{k+1}$ within a slightly non-Markovian framework. This was done to ensure that this control approach can be compared approximately with the traditional FTG algorithm that operates based on Markovian assumptions \cite{sezer2012novel}.

\subsection{Model Optimization Algorithm}
Regarding model $g(\cdot)$ (referenced in the Problem Statement \ref{sec:problem-statement}), aprior estimated next control action $\hat{y}_{T}^{-}$ is incorporated into the parameterized predictive distribution $p_{\Theta}(y_{T}|x_{T},z,R_{\Lambda},\hat{y}_{T}^{-})$. It is imperative to understand that surrogate distributions $q_{\Phi}(\cdot)$ and predictive distribution $p_{\Theta}(\cdot)$ are modeled as Gaussians due to their flexibility in approximating arbitrary distributions.

To train this model, we solve the following variational lower-bound (ELBO) optimization:
    \[ p_{\Theta}^{*}, q_{\Phi}^{*}, R_{\Lambda}^{*} = \arg\max_{\Gamma^{*}} \mathcal{L}(p_{\Theta}, q_{\Phi},R_{\Lambda}),
    \]
    subject to
\begin{align}
\label{eq:optimization}
    \text{log}\;p(y_{T}|x_{T}, \phi_{C}) &\geq \mathcal{L}(p_{\Theta},q_{\Phi},R_{\Lambda}) \nonumber \\
    &= \mathbb{E}_{z \sim q_{\Phi}}\text{log}\left[p_{\Theta}
(y_{T}|x_{T}, R_{\Lambda}, z,\hat{y}_{T}^{-})\right] \nonumber \\
       &- \mathbb{E}_{z \sim q_{\Phi}}\text{log}\left[\frac{q_{\Phi}(z|\phi_{T})}{q_{\Phi}(z|\phi_{C})}\right], 
\end{align}
%
%
where $\Gamma^{*} = \left\{\Theta^{*},\Phi^{*},\Lambda^{*}\right\}$ and the second term in the constraint equation \eqref{eq:optimization} is the KL-divergence between the approximated latent posterior $q_{\Phi}(z|\phi_{T})$ and prior $q_{\Phi}(z|\phi_{C})$.

\subsection{Speed Heuristic Utilized for FTG Algorithm} \label{subsec:speed-heuristic}

To determine the commanded speed at each timestep, we apply a simple heuristic that maps the predicted steering angle to a discrete speed value. The heuristic assigns lower speeds to higher curvature maneuvers to promote vehicle stability and safety. Specifically, steering angles exceeding a high threshold are interpreted as sharp turns, triggering a reduced "cornering" speed ($1.5\;m/s$); moderate angles correspond to a nominal "cruise" speed ($3.0\;m/s$); and near-zero steering angles—indicative of straight paths—permit the highest "fast" speed setting ($5.0\;m/s$). Explicitly, this heuristic can be mathematically represented as:
\begin{align}
    v_{k+1} =
    \begin{cases}
    1.5 \;m/s & \text{if} \; |\delta_{k+1}| > 10^{\circ} \\
    3.0\; m/s & \text{if}\; |\delta_{k+1}| > 5^{\circ} \\
    5.0\;m/s & \text{if}\; |\delta_{k+1}| \leq 5^{\circ} \\
\end{cases}
\end{align}
This rule-based mapping ensures dynamic feasibility without requiring the model to regress both speed and steering simultaneously, thereby simplifying the learning objective while retaining effective longitudinal control.

\subsection{Control Barrier Function (CBF) Formulation for Steering Safety} \label{subsec:CBF-derivation}

To guarantee safety during high-speed reactive navigation, we incorporate a control barrier function (CBF) framework that ensures obstacle avoidance using LiDAR measurements and known vehicle dynamics. The CBF is formulated as a quadratic program (QP) that minimally adjusts the neural process (NP) steering prediction while enforcing constraint satisfaction.

\subsubsection{Explicit Derivation of Vehicle Dynamics in Control-Affine Form}

We model the vehicle using a standard kinematic Ackermann steering model. Let the state of the vehicle be $x = [x, y, \theta]^\top \in \mathbb{R}^3$, where $(x, y)$ denotes the position and $\theta$ is the heading angle. The key control input is the steering angle $\delta$, giving the dynamics:
\begin{equation}
\dot{x} = f(x) + g(x) \delta =
\begin{bmatrix}
v \cos(\theta) \\
v \sin(\theta) \\
\frac{v}{L} \tan(\delta)
\end{bmatrix}
\approx
\begin{bmatrix}
v \cos(\theta) \\
v \sin(\theta) \\
\frac{v}{L} \delta
\end{bmatrix},
\end{equation}
where $v$ represents the forward velocity of the car. The approximation $\tan(\delta) \approx \delta$ is valid for small steering angles, which is reasonable under normal high-speed racing conditions where $\delta$ is bounded within $[-0.6981, 0.6981]$ radians.

\subsubsection{Safety Function Definition}

We define a measurement-based safety function $h(\psi):\mathbb{R}^{o}\rightarrow\mathbb{R}$ that depends only on the minimum observed distance between the vehicle and surrounding walls as measured by LiDAR distance array $\psi$. We denote $d_\phi$ as the minimum distance observed in the full 270 FOV (field-of-view) $o-$length LiDAR scan arc where $\phi$ is the angle within that arc that contains most dangerous (shortest) distance $d_{\phi}$. We also utilize a user-defined distance margin $d_{\text{safe}}>0$ which is set to encourage vehicle traversal ease while avoiding wall collisions ($d_{\text{safe}}\leq 0.1 m$). Then, the safety function is:
\begin{equation}
h(\psi) = d_\phi - d_{\text{safe}}
\end{equation}
Safety is ensured when $h(\psi) > 0$, i.e., the vehicle maintains a buffer of at least $d_{\text{safe}}$ meters from the nearest obstacle. 

\subsubsection{CBF Constraint Derivation}

The time derivative of the safety function is required for the CBF constraint. Since $d_\phi$ is measured in the direction of the LiDAR beam at angle $\phi$ (relative to the vehicle heading), its rate of change due to motion is:
\begin{equation}
\begin{aligned}
    \dot{h}(\psi, \delta) = \dot{d}_\phi 
    &= L_f h(\psi) + L_g h(\psi)\delta \\
    &=-v \cos(\phi) + \frac{v}{L} d_\phi \sin(\phi) \delta
\end{aligned}
\end{equation}
This defines the approximate Lie derivatives:
\begin{equation}
    \begin{aligned}
        L_f h(\psi) &= -v \cos(\phi) \\
        L_g h(\psi) &= \frac{v}{L} d_\phi \sin(\phi)
    \end{aligned}
\end{equation}
The CBF constraint enforces:
\begin{equation}
\begin{aligned}
    \dot{h}(\psi, \delta) + \alpha h(\psi) &= L_f h(\psi) + L_g h(\psi) \delta + \alpha h(\psi) \\
    &= -v\cos(\phi) +\frac{v}{L}d_{\phi}\sin(\phi)\delta +\alpha h(\psi) \\
    &\geq 0
\end{aligned}
\end{equation}
for some $\alpha > 0$ that defines the convergence rate toward the safe set.

\subsubsection{QP-Based Steering Filter}
To minimally adjust the NP-predicted steering command $\delta_{\text{raw}}$ while satisfying the CBF and steering saturation constraints, we solve the following QP at each timestep:
\begin{align}
\begin{aligned}
\delta^* = \underset{\delta \in \Delta}{\arg\min} \quad & \frac{1}{2} (\delta - \delta_{\text{raw}})^2 \\
\text{s.t} \;\; & -v\cos(\phi) +\frac{v}{L}d_{\phi}\sin(\phi)\delta + \alpha h(\psi) \geq 0 \\
& -\delta_{\max} \leq \delta \leq \delta_{\max}
\end{aligned}
\end{align}
This QP guarantees the closest safe control to the nominal NP prediction, maintaining real-time feasibility due to its single-variable convex structure.

\subsubsection{Theoretical Guarantees}

The proposed control barrier function is defined directly in the LiDAR measurement space and therefore does not admit the classical state-space forward invariance guarantees associated with exact CBF formulations \cite{ames2016control}. Instead, the QP-based steering filter provides a \emph{local, instantaneous safety guarantee} with respect to the measured environment geometry.

Specifically, at each control timestep, the solution $\delta^*$ to the QP enforces the constraint
\begin{equation}
\dot{h}(\psi,\delta^*) + \alpha h(\psi) \geq 0,
\end{equation}
which ensures that the rate of change of the measured safety margin does not decrease faster than an exponentially stabilizing bound. Thus, this condition prevents rapid degradation of the minimum observed obstacle distance. Therefore, the proposed formulation provides a \emph{practical safety certificate} that locally regulates the steering command to avoid imminent collisions and significantly reduces unsafe behaviors during high-speed reactive navigation.

\section{Results \& Discussion} \label{sec:results}
\begin{figure}
    \centering
    \includegraphics[scale=0.5]{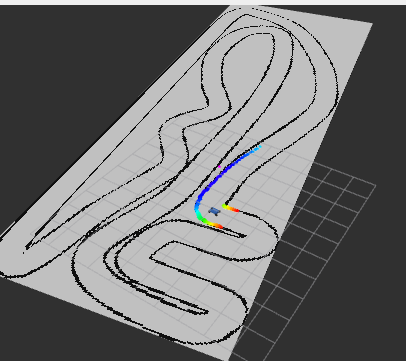}
    \caption{Screenshot of simulated F1Tenth racing environment with ego vehicle. Note that all controllers used in racing environments were low-latency ($\sim200$ hz) control methods.}
    \label{fig:racing-env}
\end{figure}
To evaluate the PI-AttNP along with other baselines in this gap-following application, these models are tasked to learn an expert FTG policy $\pi_{\text{FTG}}$ from previously-collected driving data by a FTG driver within multiple simulated F1Tenth racing environments and evaluated on an unseen racing track. An image of the utilized ego vehicle within the test racing environment can be observed in Figure \ref{fig:racing-env}. The models are first provided the required live data observed from the expert during training to ensure learning of policy $\pi_{\text{FTG}}$.

To observe how model $g(\cdot)$ is incorporated within the PI-AttNP learning structure, observe our discussion of the FTG approximation model used as $g(\cdot)$ within subsection \ref{subsec:model-used-for-PINP}. A discussion of how data from the F1tenth racing environment is structured in the context/target formulation utilized by the neural process is discussed here \ref{subsec:context-target}. The convergence behavior of the PI-AttNP and utilized baseline models is recorded and discussed in subsection \ref{subsec:converge}. Finally, a discussion of how the learned policies from PI-AttNP and other baselines perform in the autonomous racing environment in comparison to the expert FTG algorithm on the unseen-during-training racing track will be discussed \ref{subsec:race-perform}. Note that, regarding the baseline models used for benchmarking the performance of the PI-AttNP outside of the expert FTG policy, we selected the nominal Attentive Neural Process (AttNP) \cite{kim2019attentive} and the standard multi-layer perceptron with residual connections from \cite{gulino2023waymax} for improved sequential modeling (Res-MLP). See our full project implementation \href{https://github.com/devin1126/AI-Based-FTG-Reactive-Controllers}{here}.

\subsection{Approximate Control Prior for PI-AttNP} \label{subsec:model-used-for-PINP}

To augment the neural process with physically grounded inductive structure, we integrate a lightweight heuristic prior inspired by the Follow-The-Gap (FTG) algorithm. Given a forward-facing LiDAR scan $\psi\in\mathbb{R}^{o}$ with an $o-$length array of distance readings that cover the frontal 270$^{\circ}$ field-of-view of the robot, we divide the scan into $b$ equal-length angular bins. For each bin, we compute the mean range value, yielding a condensed distance bin array $\zeta \in \mathbb{R}^b$ such that $b<<o$. Let $\phi_i$ denote the center angle (in radians) of the $i$-th bin, with $\phi_1 = +\frac{3\pi}{4},+135^{\circ}$ (leftmost) and $\phi_{o} = -\frac{3\pi}{4},-135^{\circ}$ (rightmost). We identify the bin with the maximum mean distance as:
\begin{equation}
    i^* = \arg\max_i \; d_i, \quad \phi^* = \phi_{i^*}
\end{equation}
This angle $\phi^*$ serves as the direction toward the most open space and therefore acts as a proxy for the desired steering direction. Since the steering angle $\delta$ should ideally guide the vehicle toward $\phi^*$, this relationship provides an effective and interpretable approximation for a control prior:
\begin{equation}
    g(\phi_{k}) = \phi^*_{k}=\hat{y}_{T}^{-},
\end{equation}
where we only require the most recent laser scan $\psi_{k}$ as input to obtain angle $\phi^{*}_{k}$. The inclusion of $\phi^*$ in the input to the PI-AttNP decoder allows the model to condition its predictions on a strong, geometry-derived signal that correlates with steering intent. Because $\phi^*$ reflects the spatial direction of maximum free space, its sign and magnitude are typically aligned with the optimal $\delta$. Positive $\phi^*$ values indicate free space on the left (implying $\delta > 0$), while negative values indicate free space on the right (implying $\delta < 0$). During training and inference, this angle $\phi^*$ is used only as the a priori conditioning parameter of the NP decoder $\hat{y}_{T}^{-}$ to support the PI-AttNP's ability to generalize and converge efficiently across diverse navigation scenarios. Therefore, the learned PI-AttNP-based controller can still be considered an \emph{end-to-end} learned policy.

\subsection{Context/Target Formulation} \label{subsec:context-target}

In our discussion of the problem, note that the previously referenced context and target notation found in the Methodology \ref{sec:method} with $N_{C}=N_{T}=1$ describe the following:   
\begin{align*}
    x_{C} &= \left\{\zeta_{k-1},\mathcal{V}_{k-1} \right\} \in \mathbb{R}^{b+e} \\
    y_{C} &= \left\{\delta_{k} \right\} \in \mathbb{R} \\
    x_{T} &= \left\{\zeta_{k},\mathcal{V}_{k} \right \} \in \mathbb{R}^{b+e}  \\
    y_{T} &= \left\{\delta_{k+1} \right\} \in \mathbb{R},
\end{align*}

where $\zeta\in\mathbb{R}^{b}$ represents a condensed collection of equally-spaced distance bins derived from raw LiDAR scans $\psi\in\mathbb{R}^{o}$ such that $b <<o$. We also note that $\mathcal{V}\in\mathbb{R}^{e}$ represent a learned embedding vector that encodes linear/angular velocity parameters such that $\text{MLP}(v,\omega)\rightarrow\mathcal{V}$. This is done to ensure the potential input size imbalance between gap quantity $\zeta\in\mathbb{R}^{b}$ and velocity parameters $v,\omega\in\mathbb{R}^{2}$ are mitigated since both are crucial inputs in our NP-based controller. Through this, the AttNP encodes all previous dynamic information $\phi_C$ and current dynamic information in $x_{T}$ to predict desired control $y_{T}$ at time step $k+1$. 

\subsection{Real-Time Feasibility Analysis}
As can be seen in Table \ref{tab:compute-times}, we provide a table of average compute wall-time in milliseconds for all models in their respective FTG control loops along with the time for the CBF constraint:

\begin{table}[h!]
  \centering
  \scriptsize  
  \begin{tabular}{|c|c|c|c|c|}
    \hline
     & PI-AttNP FTG & AttNP FTG & Res-MLP FTG & CBF \\
    \hline
    Time (ms) & 3.414 $\pm$ 2.283 & 3.219 $\pm$ 2.318 & 3.107 $\pm$ 2.690  & 0.096 $\pm$ 0.102 \\
    \hline
    
  \end{tabular}
  \caption{Compute FTG control loop times for all models and designed CBF time}
  \label{tab:compute-times}
\end{table}

\subsection{Convergence Dynamics} \label{subsec:converge}

During the training of all deep learning-based controllers, we utilized 2000 training steps to optimize all models. Each model is trained using the Adam optimizer with learning rate $lr=1\times10^{-3}$ and batch size $|B|=100$.

\begin{figure}
    \centering
    \includegraphics[scale=0.5]{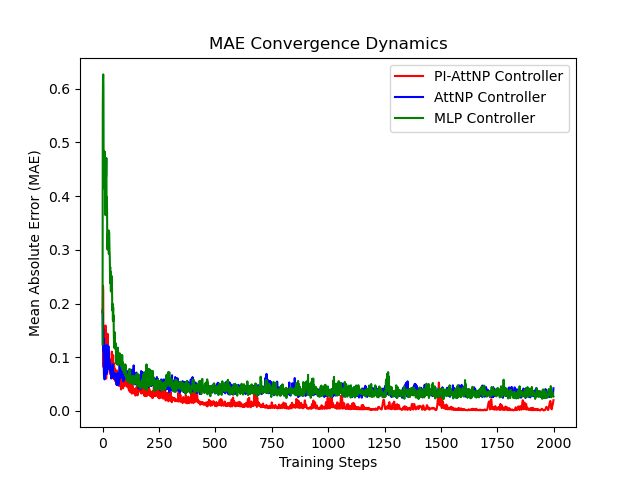}
    \caption{MAE convergence dynamics of the PI-AttNP, AttNP, and Res-MLP}
    \label{fig:mae-compare}
\end{figure}

\begin{table}[h!]
  \centering
  \scriptsize  
  \begin{tabular}{|c|c|c|c|}
    \hline
     & PI-AttNP & AttNP & Res-MLP \\
    \hline
    Lowest MAE & 0.00032 & 0.02005 & 0.01719 \\
    \hline
    
  \end{tabular}
  \caption{Lowest MAE reached for all models in Figure \ref{fig:mae-compare}}
  \label{tab:train-MAE-results}
\end{table}

\begin{figure}
    \centering
    \includegraphics[scale=0.5]{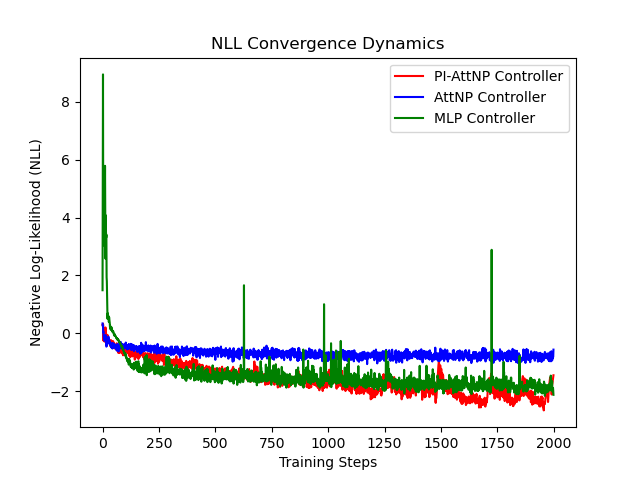}
    \caption{Distribution error dynamics of the PI-AttNP, AttNP, and Res-MLP}
    \label{fig:nll-compare}
\end{figure}

\begin{table}[h!]
  \centering
  \scriptsize  
  \begin{tabular}{|c|c|c|c|}
    \hline
     & PI-AttNP & AttNP & Res-MLP\\
    \hline
    Lowest NLL & -2.6600 & -1.06039 & -2.15674 \\
    \hline
  \end{tabular}
  \caption{Lowest NLL reached for all models in Figure \ref{fig:nll-compare}}
  \label{tab:train-NLL-results}
\end{table}

Figure \ref{tab:train-MAE-results} and Figure \ref{tab:train-NLL-results} show the convergence behavior of each learned controller under identical optimization settings. Across the full 2000 training steps, the PI-AttNP consistently achieves the lowest mean absolute error (MAE) over the AttNP and Res-MLP baselines. In terms of distribution quality, PI-AttNP also attains the lowest negative log-likelihood (NLL) in comparison to both baselines. These results indicate that injecting the approximate FTG-based prior into the NP decoder improves both predictive accuracy and uncertainty calibration.

\subsection{Racing Performance Analysis} \label{subsec:race-perform}

\begin{table}[h!]
  \centering
  \scriptsize 
  \begin{tabular}{|c|c|c|c|}
    \hline
    \rule{0pt}{2.6ex} Models & Avg TTF $(s)$ & Avg $\#$ of Collisions & Avg $\dot{\delta}$ ($\text{rad}/s$) \\
    \hline
    FTG (Expert) & 40.3 & 0.4 & 0.010288 \\
    \hline
    PI-AttNP & 40.4 & 0.4 & 0.004262 \\
    \hline
    PI-AttNP (CBF) & 40.6 & 0.2 & 0.009561 \\
    \hline
    AttNP & 40.1 & 0.8 & 0.005566 \\
    \hline 
    Res-MLP & 39.5 & 1.2 & 0.016821 \\
    \hline
  \end{tabular}
  \caption{Performance of all models in racing environment after five recorded laps}
  \label{tab:racing-perform}
\end{table}

As can be observed in Table \ref{tab:racing-perform}, we evaluate closed-loop performance of each controller for five recorded laps on an unseen racing track and assessed using average time-to-finish (TTF), collision count, and average steering rate $\dot{\delta}$. All models achieve comparable lap times near 40 seconds, showing that the environment can be completed at similar speeds when stable behavior is maintained. However, clear safety differences emerge: the PI-AttNP (CBF) achieves the lowest collision rate at 0.2 per run while preserving competitive TTF, demonstrating the effectiveness of the QP-based safety filter in reducing impacts without heavily sacrificing performance. In contrast, the AttNP and Res-MLP exhibit increased collision frequency (0.8 and 1.2, respectively), suggesting that purely learned reactive policies remain more susceptible to unsafe cornering and wall interactions even when trained to imitate FTG. Overall, PI-AttNP provides the most favorable balance between imitation accuracy, stability, and safety, while the CBF further improves reliability in high-speed navigation.

\section{Conclusion \& Future Works} \label{sec:conclude+future}
This paper introduced a neural process-based reactive control framework for autonomous racing using an Attentive Neural Process (AttNP) and a model-informed extension (PI-AttNP) that incorporates an approximate FTG-based control prior. Experimental results demonstrate that PI-AttNP achieves substantially improved convergence behavior over purely learned baselines, producing lower prediction error and better distributional performance during training. In closed-loop racing evaluation, PI-AttNP matches expert FTG lap completion performance while reducing collisions relative to AttNP and Res-MLP, and the addition of a CBF-based steering filter further improves safety by enforcing a real-time constraint on the measured obstacle clearance. Future work will explore longer-horizon context histories for autoregressive policies, multi-step rollout prediction for model predictive control, and richer safety formulations that account for explicit vehicle states and stability certificates.

\bibliographystyle{unsrt}
\bibliography{sources-attnp-ftg}

@article{sezer2012novel,
  title={A novel obstacle avoidance algorithm:“Follow the Gap Method”},
  author={Sezer, Volkan and Gokasan, Metin},
  journal={Robotics and Autonomous Systems},
  volume={60},
  number={9},
  pages={1123--1134},
  year={2012},
  publisher={Elsevier}
}

@article{borenstein1991vector,
  title={The vector field histogram-fast obstacle avoidance for mobile robots},
  author={Borenstein, Johann and Koren, Yoram and others},
  journal={IEEE transactions on robotics and automation},
  volume={7},
  number={3},
  pages={278--288},
  year={1991}
}

@article{khatib1986real,
  title={Real-time obstacle avoidance for manipulators and mobile robots},
  author={Khatib, Oussama},
  journal={The international journal of robotics research},
  volume={5},
  number={1},
  pages={90--98},
  year={1986},
  publisher={Sage Publications Sage CA: Thousand Oaks, CA}
}

@inproceedings{koren1991potential,
  title={Potential field methods and their inherent limitations for mobile robot navigation.},
  author={Koren, Yoram and Borenstein, Johann and others},
  booktitle={Icra},
  volume={2},
  number={1991},
  pages={1398--1404},
  year={1991}
}

@article{bojarski2016end,
  title={End to end learning for self-driving cars},
  author={Bojarski, Mariusz and Del Testa, Davide and Dworakowski, Daniel and Firner, Bernhard and Flepp, Beat and Goyal, Prasoon and Jackel, Lawrence D and Monfort, Mathew and Muller, Urs and Zhang, Jiakai and others},
  journal={arXiv preprint arXiv:1604.07316},
  year={2016}
}

@article{kendall2017uncertainties,
  title={What uncertainties do we need in bayesian deep learning for computer vision?},
  author={Kendall, Alex and Gal, Yarin},
  journal={Advances in neural information processing systems},
  volume={30},
  year={2017}
}

@article{kiran2021deep,
  title={Deep reinforcement learning for autonomous driving: A survey},
  author={Kiran, B Ravi and Sobh, Ibrahim and Talpaert, Victor and Mannion, Patrick and Al Sallab, Ahmad A and Yogamani, Senthil and Perez, Patrick},
  journal={IEEE transactions on intelligent transportation systems},
  volume={23},
  number={6},
  pages={4909--4926},
  year={2021},
  publisher={IEEE}
}

@article{ames2016control,
  title={Control barrier function based quadratic programs for safety critical systems},
  author={Ames, Aaron D and Xu, Xiangru and Grizzle, Jessy W and Tabuada, Paulo},
  journal={IEEE Transactions on Automatic Control},
  volume={62},
  number={8},
  pages={3861--3876},
  year={2016},
  publisher={IEEE}
}

@inproceedings{sun2021learning,
  title={Learning certified control using contraction metric},
  author={Sun, Dawei and Jha, Susmit and Fan, Chuchu},
  booktitle={Conference on Robot Learning},
  pages={1519--1539},
  year={2021},
  organization={PMLR}
}

@article{berkenkamp2017safe,
  title={Safe model-based reinforcement learning with stability guarantees},
  author={Berkenkamp, Felix and Turchetta, Matteo and Schoellig, Angela and Krause, Andreas},
  journal={Advances in neural information processing systems},
  volume={30},
  year={2017}
}

@misc{dubois2020npf,
  title        = {Neural Process Family},
  author       = {Dubois, Yann and Gordon, Jonathan and Foong, Andrew YK},
  month        = {September},
  year         = {2020},
  howpublished = {\url{http://yanndubs.github.io/Neural-Process-Family/}}
}

@article{kim2019attentive,
  title={Attentive neural processes},
  author={Kim, Hyunjik and Mnih, Andriy and Schwarz, Jonathan and Garnelo, Marta and Eslami, Ali and Rosenbaum, Dan and Vinyals, Oriol and Teh, Yee Whye},
  journal={arXiv preprint arXiv:1901.05761},
  year={2019}
}

@article{gulino2023waymax,
  title={Waymax: An accelerated, data-driven simulator for large-scale autonomous driving research},
  author={Gulino, Cole and Fu, Justin and Luo, Wenjie and Tucker, George and Bronstein, Eli and Lu, Yiren and Harb, Jean and Pan, Xinlei and Wang, Yan and Chen, Xiangyu and others},
  journal={Advances in Neural Information Processing Systems},
  volume={36},
  pages={7730--7742},
  year={2023}
}

\end{document}